\documentclass[10pt,twocolumn,letterpaper]{article}

\usepackage{iccv}
\usepackage{times}
\usepackage{epsfig}
\usepackage{graphicx}
\usepackage{amsmath}
\usepackage{amssymb}

\usepackage{booktabs}
\usepackage{multirow}
\usepackage{subcaption}
\usepackage[table]{xcolor}
\usepackage{bbm}
\usepackage{enumitem}

\usepackage[accsupp]{axessibility}  

\newcommand{\parsection}[1]{\noindent\textbf{#1:}~}

\definecolor{baselinecolor}{gray}{.9}
\newcommand{\baseline}[1]{\cellcolor{baselinecolor}{#1}}

\definecolor{citecolor2}{HTML}{0071bc}
\usepackage[pagebackref=true,breaklinks=true,letterpaper=true,colorlinks,urlcolor=red,citecolor=blue,linkcolor=red,bookmarks=false]{hyperref}

\iccvfinalcopy 

\def\iccvPaperID{2063} 

\ificcvfinal\fi

\begin{document}

\title{Masked Motion Predictors are Strong 3D Action Representation Learners}

\author{Yunyao~Mao$^{1}$~~~~Jiajun~Deng$^{3}$~~~~Wengang~Zhou$^{1,2,}$\thanks{Corresponding authors: Wengang Zhou and Houqiang Li}~~~~Yao~Fang$^{4}$~~~~Wanli~Ouyang$^{3}$~~~~Houqiang~Li$^{1,2,*}$ \\
	{\normalsize $^{1}$CAS Key Laboratory of Technology in GIPAS, EEIS Department, University of Science and Technology of China} \\
	{\normalsize $^{2}$Institute of Artificial Intelligence, Hefei Comprehensive National Science Center} \\
    {\normalsize $^{3}$The University of Sydney}~~~
    {\normalsize $^{4}$Merchants Union Consumer Finance Company Limited} \\
	{\tt\small myy2016@mail.ustc.edu.cn, jiajun.deng@sydney.edu.au, zhwg@ustc.edu.cn} \\{\tt\small fangyao@mucfc.com, wanli.ouyang@sydney.edu.au, lihq@ustc.edu.cn}
}

\maketitle
\ificcvfinal\thispagestyle{empty}\fi

\begin{abstract}
In 3D human action recognition, limited supervised data makes it challenging to fully tap into the modeling potential of powerful networks such as transformers. As a result, researchers have been actively investigating effective self-supervised pre-training strategies. In this work, we show that instead of following the prevalent pretext task to perform masked self-component reconstruction in human joints, explicit contextual motion modeling is key to the success of learning effective feature representation for 3D action recognition. Formally, we propose the \underline{Ma}sked \underline{M}otion \underline{P}rediction (MAMP) framework. To be specific, the proposed MAMP takes as input the masked spatio-temporal skeleton sequence and predicts the corresponding temporal motion of the masked human joints. Considering the high temporal redundancy of the skeleton sequence, in our MAMP, the motion information also acts as an empirical semantic richness prior that guide the masking process, promoting better attention to semantically rich temporal regions. Extensive experiments on NTU-60, NTU-120, and PKU-MMD datasets show that the proposed MAMP pre-training substantially improves the performance of the adopted vanilla transformer, achieving state-of-the-art results without bells and whistles. The source code of our MAMP is available at \url{https://github.com/maoyunyao/MAMP}.
\end{abstract}

\section{Introduction}
\label{sec:intro}

\begin{figure}[t]
\centering
\begin{minipage}{0.9\linewidth}
	\centering
	\includegraphics[width=1.0\linewidth]{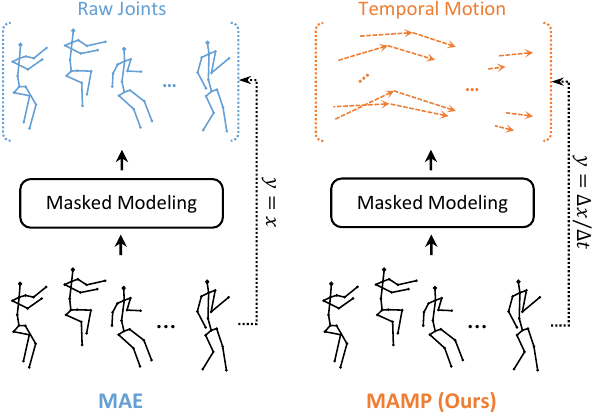}
	\subcaption{Comparison of pre-training objectives.}
	\vspace{2mm}
\end{minipage}
\begin{minipage}{0.9\linewidth} 
	\centering
	\includegraphics[width=1.0\linewidth]{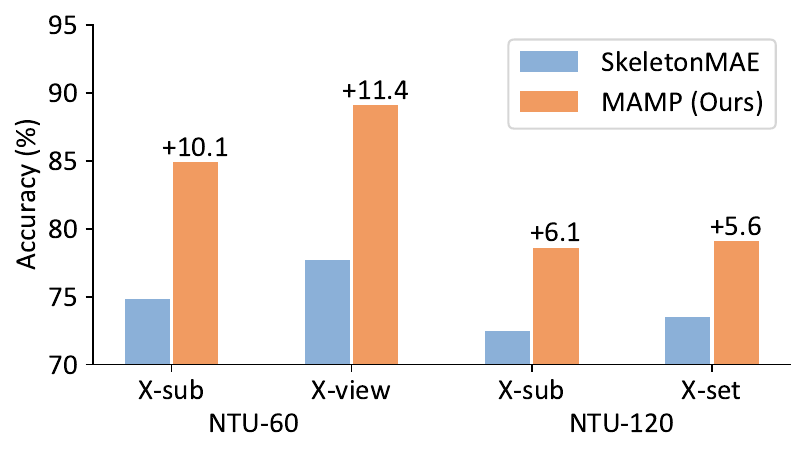}
	\subcaption{Comparison of linear probing accuracy.}
\end{minipage}
\vspace{-2mm}
\caption{
Illustration of (a) pre-training objective comparison between masked auto encoders (MAE) and our masked motion predictors (MAMP) and (b) performance comparison between the typical MAE method, \emph{i.e.}, SkeletonMAE \cite{skeletonmae}, and our MAMP under the linear evaluation protocol.}
\label{fig:intro}
\end{figure}

How to accurately recognize human actions has been a long-standing challenge in computer vision. Recently, with the advances in techniques of depth sensing and pose estimation~\cite{openpose,fang2017rmpe,xu2020deep}, skeleton-based 3D human action recognition has become an emerging problem to the community, which is of great significance in a series of applications such as human-computer interaction, video surveillance, virtual reality, \emph{etc}.
Despite the computation efficiency and background robustness of skeletons, existing supervised 3D action recognition methods \cite{Chen_2021_ICCV,cheng2020skeleton,du2015hierarchical,ke2017new,li2019actional,Li_2021_ICCV,liu2020disentangling,shi2019skeleton,Shi_2021_ICCV,si2019attention,zhang2020semantics} heavily rely on well-annotated training sequences, which are labor-intensive and time-consuming to acquire. Furthermore, limited supervision also leads to the overfitting issue in general models, especially for transformers that are with weak inductive bias and high model capacity. 
These facts motivate the exploration of self-supervised 3D action representation learning.

In the literature, the prevalent pretext tasks originally developed for images have been adapted for 3D action representation learning, such as colorization \cite{yang2021skeleton}, reconstruction \cite{zheng2018unsupervised,su2020predict,lin_ms2l_mm}, contrastive learning \cite{li20213d,thoker2021skeleton,guo2022aimclr}, \emph{etc}. Among them, contrastive learning once dominated 3D action representation learning with its concise framework and promising performance. Nevertheless, as a global representation learner, it still suffers from certain limitations, such as the lack of explicit constraints for temporal context modeling and the over-reliance on heuristic action data augmentations \cite{liu2021self}, impeding its further exploration of 3D actions.



Recently, as transformers flourish in computer vision, masked autoencoder (MAE) \cite{he2022masked} has attracted a surge of research interest for its exceptional performance. Given that a 3D skeleton serves as an abstract representation of human behaviors, there has been growing interest in applying the MAE concept to 3D action representation learning, to capture the underlying spatio-temporal dynamics of skeleton sequences. Early attempts generally followed the practice of images, employing masked self-reconstruction of human joints as the pre-training pretext. Despite considerable effort, we argue that the network is not effectively directed to prioritize contextual motion modeling in such a self-reconstruction objective, which is, however, crucial for comprehending 3D actions as the appearance information is greatly erased in human skeletons. 
How to better explore the contextual motion clue in self-supervised 3D action representation learning is a valid problem.


By consolidating this idea, we introduce \underline{Ma}sked \underline{M}otion \underline{P}rediction (MAMP), a simple yet effective framework to address the problem of self-supervised 3D action representation learning. Specifically, the proposed MAMP takes as input the masked spatio-temporal skeleton sequence and turns to predict the corresponding temporal motion of the masked human joints. In this way, the network is directly encouraged for contextual motion modeling. Moreover, given the observation that moments with significant motion are often critical for human action understanding, in our MAMP, the temporal motion is used not only as the pre-training objective but also as an empirical semantic richness prior that effectively guiding the skeleton masking process. Compared to the random version, the proposed motion-aware masking strategy takes additional temporal motion intensity as input. It first converts the input intensity into a probability distribution and then utilizes the re-parameterization technique for efficient probability-guided masked token sampling. As a result, joints with significant motion are masked with a higher probability, facilitating better attention to semantically rich temporal regions.

As illustrated in Figure \ref{fig:intro}, compared to masked self-reconstruction of human joints, masked motion prediction acts as a more effective pretext task for 3D action representation learning. It substantially alleviates the problem that the transformers cannot fully unleash their modeling potential for human actions due to the scarcity of annotated 3D skeletons. The adopted vanilla transformer sets a series of state-of-the-art records in 3D action recognition after MAMP pre-training, without the need for bells and whistles such as multi-stream ensembling. Specifically, compared to training from scratch, our MAMP demonstrates significant absolute performance improvements of 10.0\% and 13.2\% on the challenging cross-subject protocol of NTU RGB+D 60 \cite{shahroudy2016ntu} and NTU RGB+D 120 \cite{liu2020ntu} datasets, resulting in top-1 accuracy of 93.1\% and 90.0\%, respectively. We hope this simple yet effective framework will serve as a strong baseline that facilitates future research on 3D action pre-training and beyond.

Overall, we make the following three-fold contributions:
\begin{itemize}[noitemsep,nolistsep]
\item We present masked motion prediction to learn 3D action representation, which substantially alleviates the insufficient contextual motion modeling issue in the conventional masked self-reconstruction paradigm.
\item We devise the motion-aware masking strategy, which incorporates motion intensity as an empirical semantic richness prior for adaptive joint masking.
\item We conduct extensive experiments on three prevalent benchmarks to verify the effectiveness of our method. Remarkably, with our proposed MAMP, the vanilla transformer, for the first time, achieves the top-performing record for 3D action recognition.

\end{itemize}


\section{Related Work}

\subsection{Supervised 3D Action Recognition}
How to better model the dynamic skeletons for supervised action recognition is an extensively studied problem. In many early works, RNNs are favored for their excellent sequential modeling capability, such as the hierarchical RNN model proposed in \cite{du2015hierarchical} and the 2D Spatio-Temporal LSTM in \cite{liu2016spatio,liu2017skeleton}. In view of the great success of CNNs \cite{NIPS2012_alexnet,he2016deep} in image understanding, some methods also try to apply it to 3D action recognition. To cater for the input format, \cite{YongDu2015SkeletonBA} and \cite{li2017skeleton} treat the skeleton sequence as a three-channel (x, y, and z coordinates) pseudo-image, with the number of frames and joints as height and width, respectively. Considering the natural connections between joints, ST-GCN \cite{yan2018spatial} introduces the Graph Neural Networks (GCNs) for skeleton modeling, where the convolution kernels are elaborately designed according to the skeleton topology. The astonishing performance of ST-GCN has led the trend of GCN-based 3D action recognition, with numerous subsequent improvements emerging in input streams \cite{wang2017modeling,liang2019three,2sagcn2019cvpr}, kernel design \cite{Chen_2021_ICCV,zhang2020context,shi2019skeleton,liu2020disentangling}, \emph{etc}.

Recent approaches \cite{qiu2022spatio,plizzari2021skeleton,dstanet_accv2020} try to introduce the popular vision transformer into 3D action recognition. However, under limited training data, vanilla transformers with weak inductive bias cannot be fully trained. Therefore, many customized designs are required in existing supervised attempts, such as temporal convolution \cite{qiu2022spatio}, graph convolution \cite{plizzari2021skeleton,qiu2022spatio}, space-time separation \cite{dstanet_accv2020}, \emph{etc}. In our approach, we demonstrate that pre-training with masked motion prediction is key to the success of transformers in 3D action recognition. The proposed MAMP framework endows the vanilla transformer with unrivaled performance.

\subsection{Self-supervised 3D Action Recognition}
Self-supervised representation learning aims to capture the domain priors from unlabeled data so as to facilitate the application of the model in downstream tasks. In 3D human action recognition, many pretext tasks have been utilized to explore the action context that resides in the skeleton sequence. Among them, LongT GAN \cite{zheng2018unsupervised} and P\&C \cite{su2020predict} try to learn 3D action representation by autoencoder-based sequence reconstruction, where the decoder in P\&C is further weakened to promote the learning of the feature encoder. In Colorization \cite{yang2021skeleton}, the skeleton sequences are treated as point clouds and action representation is learned by colorizing each joint based on its spatial and temporal orders.

Recently, many contrastive learning-based approaches \cite{lin_ms2l_mm,thoker2021skeleton,li20213d,guo2022aimclr,cpm,cmd} have emerged, showing superior performance compared to earlier works. To learn better 3D action representation, they either try to dig helpful supervision across different skeleton modalities \cite{li20213d,cmd}, or explore better action data augmentation \cite{guo2022aimclr} and positive sample mining strategies \cite{cpm}. Nevertheless, as a global feature learner originally designed for images, contrastive learning lacks explicit constraints on the exploration of temporal motion context, limiting its further development for 3D actions.

SkeletonMAE \cite{skeletonmae} first introduces the idea of MAE\cite{he2022masked} into transformer-based 3D action representation learning, where the original joint coordinates of masked regions are predicted. In our approach, we demonstrate that such a self-reconstruction objective is sub-optimal for learning 3D action representation. Therefore, we introduce the Masked Motion Prediction (MAMP) framework for explicit contextual motion modeling, resulting in significantly better performance compared to raw skeleton reconstruction.

\subsection{Masked Visual Prediction}
With the development of vision transformers \cite{carion2020end,dosovitskiy2020vit,pmlr-v139-touvron21a}, the masked prediction derived from the autoencoder \cite{ballard1987modular} has revived again. Similar to the BERT \cite{bert} pre-training in NLP, the input tokens are randomly masked and corresponding objectives are predicted, which can be the raw pixels \cite{he2022masked}, HOG features \cite{wei2022masked}, or token ids from offline learned dVAEs \cite{bao2022beit}. Recently, there have also been attempts \cite{arxiv.2210.05234,arxiv.2210.04154} to use optical flow or temporal difference of images as the auxiliary reconstruction objectives, but inferior performance is observed when they are applied alone. This is largely attributed to the high redundancy of the raw images, where the key foreground motion is difficult to be pre-extracted accurately. In our approach, we employ the idea of masked visual prediction for 3D action representation learning, with the temporal skeleton motion adopted as the only reconstruction target. Different from images, the explicit temporal correspondence of joints in the human skeleton sequence enables the ready extraction of their accurate motion context. Furthermore, we also incorporate motion intensity as the semantic richness prior to guide the masking process.

\begin{figure*}[t!]
    \centering
    \includegraphics[width=1.0\linewidth]{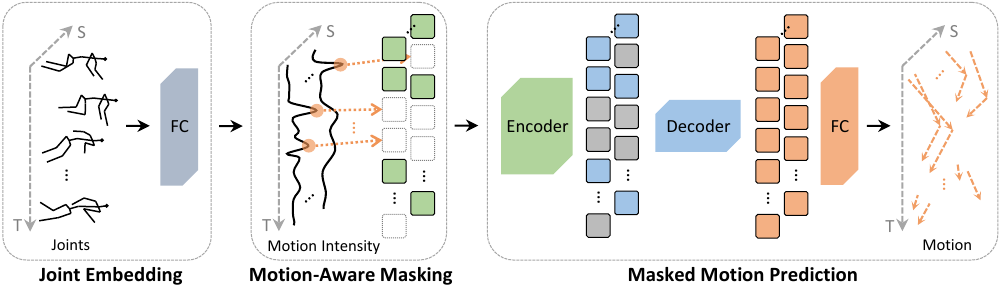}
    \vspace{-6mm}
    \caption{The overall pipeline of the proposed MAMP framework. Different from the self-reconstruction scheme adopted in previous works \cite{zheng2018unsupervised,skeletonmae}, our MAMP learns 3D action representation by predicting the corresponding motion sequence for masked joint input. Moreover, motion information also acts as an empirical semantic richness prior that effectively guides the masking process, enabling more attention to be applied to regions with significant temporal motion intensity.}
    \label{fig:arch}
\end{figure*}

\section{Our Method}

\subsection{Overview}

Figure \ref{fig:arch} illustrates the overall pipeline of our proposed \underline{Ma}sked \underline{M}otion \underline{P}rediction (MAMP) framework. It takes a skeleton sequence $S \in \mathbb{R}^{T_s \times V \times C_s}$ as input, which is randomly cropped from the original data and is resized to a fixed temporal length $T_s$. $V$ and $C_s$ are the number of joints and coordinate channels, respectively. The motion sequence $M \in \mathbb{R}^{T_s \times V \times C_s}$ of the input is also extracted, which is defined as the differential on temporal dimension (manually padded for the first frame).

As in most vision transformers, the input joints are linearly mapped into joint embedding $E \in \mathbb{R}^{T_e \times V \times C_e}$. After that, the motion-aware masking strategy is applied to mask most of the embedding features under the guidance of temporal motion intensity. The remaining features are processed by the encoder-decoder architecture, where the transformer encoder learns representation from unmasked joint embedding and the transformer decoder performs contextual modeling based on the learnable mask tokens and latent representation from the transformer encoder. Different from MAE \cite{he2022masked} that reconstructs the original signal for representation learning, in MAMP, a motion prediction head is adopted, which takes decoded features as input and predicts the temporal motion of the input skeleton sequence.

After the aforementioned pre-training, only the joint embedding layer and the transformer encoder are reserved for downstream applications.

\subsection{Joint Embedding}
In most transformer-based attempts \cite{plizzari2021skeleton,dstanet_accv2020,qiu2022spatio}, each spatio-temporal skeleton joint is embedded separately, resulting in a large number of input tokens. Considering the temporal redundancy, in our approach, the input skeleton sequence $S \in \mathbb{R}^{T_s \times V \times C_s}$ is divided into temporally non-overlapping segments $S' \in \mathbb{R}^{T_e \times V \times l \times C_s}$, where $l$ is the length of each segment and $T_e = T_s / l$. In each segment, joints with the same spatial position are embedded together:
\begin{equation}
  E = \texttt{JointEmbed}(S') \in \mathbb{R}^{T_e \times V \times C_e},
\end{equation}
where $C_e$ is the dimension of the embedding features. Compared to the original skeleton sequence, the temporal resolution of the embedding $E$ is reduced by a factor of $l$, resulting in higher computational efficiency.

\subsection{Motion Extraction}
Different from RGB frames with heavy spatial redundancy, the human skeleton sequence is highly semantic, with explicit correspondence between neighboring frames. Therefore, we can easily obtain its motion $M \in \mathbb{R}^{T_s \times V \times C_s}$ by applying temporal difference on joint coordinates:
\begin{equation}
    \label{eq:motion}
    \begin{aligned}
    M_{i,:,:} = S_{i,:,:} - S_{i-m,:,:}, \,\, i \in {m,m+1,\dots, T_s-1}, \\
    \end{aligned}
\end{equation}
where stride $m$ controls the step size of the motion. For convenience, the motion sequence is padded to be consistent with the length of the original input:
\begin{equation}
    \label{eq:motion_padding}
    M_{0:m-1,:,:} = \left\{
    \begin{array}{ll}
    0, & {constant} \\
    M_{m:2m-1,:,:}, & {replicate}
    \end{array}
    \right.
    ,
\end{equation}
where ${constant}$ and ${replicate}$ denote constant padding (with zeros) and replicate padding for the first $m$ frames of the motion sequence, respectively.
\subsection{Motion-Aware Masking}
In the proposed approach, the motion information is used not only as the reconstruction target during pre-training but also as the empirical semantic richness prior that guiding the masking of embedding features. Considering that the skeleton sequence is segment-wise embedded with length $l$, we extract the motion sequence $M^{\text{mask}} \in \mathbb{R}^{T_i \times V \times C_i}$ with stride $m = l$ and replicate padding according to Eq. \eqref{eq:motion} and Eq. \eqref{eq:motion_padding}, which is further reshaped into $M' \in \mathbb{R}^{T_e \times V \times l \times C_i}$ as is done for $S'$. Then, the motion intensity $I$, which indicates the motion significance of each spatio-temporal segment, is computed as follows:
\begin{equation}
  I = \sum_{i=0}^{l}\sum_{j=0}^{C_i} |M'_{:,:,i,j}| \in \mathbb{R}^{T_e \times V}.
\end{equation}

Since human actions are composed of a series of temporal movements, we argue that the intensity of motion largely reflects the semantic richness. Therefore in MAMP, the motion intensity is further converted into probability distribution with a temperature hyper-parameter $\tau$:
\begin{equation}
  \pi = \texttt{Softmax}(I / \tau),
\end{equation}
which indicates the probability that each embedding feature is masked. In MAMP, the idea of gumble max is adapted for efficient probability-guided mask index sampling:
\begin{equation}
    \label{eq:idx_mask}
    \begin{aligned}
    g &= - \log ( - \log \epsilon), \, \epsilon \in U[0,1]^{T_e \times V}, \\
    idx^{\text{mask}} &= \texttt{Index-of-Top-K} (\log \pi + g), \\
    \end{aligned}
\end{equation}
where $U[0,1]$ denotes uniform distribution between 0 and 1. The obtained $idx^{\text{mask}}$ indicates which joints are masked and is used for unmasked token selection in Section \ref{section:mamp}. Based on the above operations, the network is encouraged to focus more on semantically rich regions, so as to learn more discriminative 3D action representation.

\subsection{Masked Motion Prediction}
\label{section:mamp}
We follow the encoder-decoder design in MAE \cite{he2022masked}, where the transformer encoder focuses on representation learning, while the decoder is responsible for the implementation of the pre-training pretext.

\parsection{Encoder} In the encoder, separate spatio-temporal positional embedding $P_e^s \in \mathbb{R}^{1 \times V \times C_e}$ and $P_e^t \in \mathbb{R}^{T_e \times 1 \times C_e}$ are first element-wise added (with broadcasting) to the input joint embedding $E$:
\begin{equation}
  \begin{aligned}
    E_p = E + P_e^s + P_e^t. \\
  \end{aligned}
\end{equation}
Then, the unmasked tokens in $E_p$ are selected according to the $idx^{\text{mask}}$ extracted in Eq. \eqref{eq:idx_mask} and are flattened to $E_p^u \in \mathbb{R}^{N_u \times C_e}$, where $N_u = T_e \times V \times (1 - \text{mask ratio})$ is the number of unmasked tokens. After that, the latent representation is extracted by $L_e$ vanilla transformer blocks:
\begin{equation}
  \begin{aligned}
    H_0 &= E_p^u, \\
    H_l' &= \texttt{MSA}(\texttt{LN}(H_{l-1})) + H_{l-1}, &l \in 1, \cdots, L_e \\
    H_l &= \texttt{MLP}(\texttt{LN}(H_l')) + H_l',  &l \in 1, \cdots, L_e \\
    H_e^u &= \texttt{LN}(H_{L_e}),
  \end{aligned}
\end{equation}
where $\texttt{MSA}$, $\texttt{MLP}$, and $\texttt{LN}$ denote multi-head self-attention, multilayer perceptron, and layer norm, respectively.

\parsection{Decoder} In the decoder, the learnable mask tokens are inserted into $H_e^u$ according to the mask indices $idx^{\text{mask}}$. The result is reshaped back to $H_e \in \mathbb{R}^{T_e \times V \times C_e}$, which is processed by $L_d$ decoder layers for masked modeling:
\begin{equation}
  \begin{aligned}
    Z_0 &= H_e + P_d^s + P_d^t, \\
    Z_l' &= \texttt{MSA}(\texttt{LN}(Z_{l-1})) + Z_{l-1}, &l \in 1, \cdots, L_d \\
    Z_l &= \texttt{MLP}(\texttt{LN}(Z_l')) + Z_l',  &l \in 1, \cdots, L_d \\
    Z_d &= \texttt{LN}(Z_{L_d}),
  \end{aligned}
\end{equation}
where $P_d^s$ and $P_d^t$ are the spatial and temporal positional embedding of the transformer decoder, respectively.

\parsection{Motion Prediction}
In our method, the reconstruction target is not the original skeletons, but the motion sequence $M^{\text{target}}$ pre-extracted according to Eq. \eqref{eq:motion} and Eq. \eqref{eq:motion_padding}, which is normalized by its segment-wise mean and standard deviation as in \cite{he2022masked}. Therefore, given the decoded feature $Z_d \in \mathbb{R}^{T_e \times V \times C_d}$, we additionally adopt a prediction head to predict the temporal motion of masked human joints:
\begin{equation}
  M^{\text{pred}} = \texttt{MotionPredHead}(Z_d),
\end{equation}
where we empirically find that a simple fully connected layer just works well. For masked human joints, we compute the mean squared error (MSE) between the predicted result $M^{\text{pred}}$ and the reconstruction target $M^{\text{target}}$:
\begin{equation}
  \begin{aligned}
    \mathcal{L} = \frac{1}{|idx^{\text{mask}}|} \sum_{(i,j) \in idx^{\text{mask}}}\|(M_{i,j,:}^{\text{pred}} - M_{i,j,:}^{\text{target}})\|_{2}^{2}.
  \end{aligned}
\end{equation}

\section{Experiments}

\subsection{Datasets and Evaluation Protocols}
\parsection{NTU-RGB+D 60 \cite{shahroudy2016ntu}} NTU-RGB+D 60 (NTU-60) is a large-scale dataset for human action recognition. It contains 60 action categories performed by 40 different subjects, with a total of 56,880 3D skeleton sequences. In this paper, we adopt the evaluation protocols recommended by the authors, namely cross-subject (X-sub) and cross-view (X-view). The former, X-sub uses the action sequences performed by half of the 40 subjects as training samples and the rest as test samples. For X-view, the training samples are action sequences captured by cameras 2 and 3, and the test samples are those captured by camera 1.

\parsection{NTU-RGB+D 120 \cite{liu2020ntu}} NTU-RGB+D 120 (NTU120) is an extended version of NTU-60, in which the number of action categories is increased from 60 to 120, the number of total skeleton sequences and subjects are also increased to 114,480 and 106, respectively. Furthermore, the authors also introduce a more challenging evaluation protocol named cross-setup (X-set) to substitute for the original X-view in NTU-60. Specifically, X-set divides sequences into 32 different setups based on the camera distance and background. Samples from half of these setups are used as the training set and the remainder constitute the test set.

\parsection{PKU-MMD \cite{liu2017pku}} Following \cite{thoker2021skeleton}, to perform 3D action classification on PKU-MMD, we crop out action instances based on temporal annotations and divide them into training and test sets according to the cross-subject protocol. PKU-MMD contains two phases: PKU-MMD I (PKU-I) and PKU-MMD II (PKU-II). In PKU-I, the number of samples in training and test sets are 18,841 and 2,704, respectively. Due to the more noise introduced by the larger view variation, PKU-II is more challenging, with 5,332 samples for training and 1,613 for testing.


\begin{table*}[t!]
	\centering
    \setlength\tabcolsep{10.2pt}
    \begin{tabular}{lccccccc}
      \toprule[1.0pt]
      \multirow{2}{*}{Method} & \multirow{2}{*}{Input stream} &  \multicolumn{2}{c}{\textbf{NTU-60}}	& \multicolumn{2}{c}{\textbf{NTU-120}} & \multicolumn{2}{c}{\textbf{PKU-MMD}} \\
      \cmidrule[0.4pt](lr){3-4} \cmidrule[0.4pt](lr){5-6} \cmidrule[0.4pt](lr){7-8}
                                                & & X-sub & X-view & X-sub & X-set & Phase I & Phase II \\
      \midrule[0.4pt]
      3s-SkeletonCLR \cite{li20213d}            & Joint+Motion+Bone & 75.0  & 79.8	 & 60.7  & 62.6 & 85.3 & - \\
      3s-CrosSCLR \cite{li20213d}               & Joint+Motion+Bone & 77.8  & 83.4	 & 67.9  & 66.7 & 84.9 & 21.2 \\
      3s-AimCLR \cite{guo2022aimclr}	        & Joint+Motion+Bone & 78.9  & 83.8   & 68.2  & 68.8 & 87.4 & 39.5 \\
      \midrule[0.4pt]
      LongT GAN \cite{zheng2018unsupervised}    & Joint only & 39.1  & 48.1   & -     & -    & 67.7 & 26.0 \\
      P\&C \cite{su2020predict}		            & Joint only & 50.7  & 76.3   & 42.7  & 41.7 & 59.9 & 25.5 \\
      MS$^2$L \cite{lin_ms2l_mm}                & Joint only & 52.6  & -      & -     & -    & 64.9 & 27.6 \\
      AS-CAL \cite{rao2021augmented} 			& Joint only & 58.5  & 64.8   & 48.6  & 49.2 & - & - \\
      ISC \cite{thoker2021skeleton}		        & Joint only & 76.3  & 85.2   & 67.1  & 67.9 & 80.9 & 36.0 \\
      GL-Transformer \cite{gl_transformer}      & Joint only & 76.3  & 83.8   & 66.0  & 68.7 & -    & - \\
      CPM  \cite{cpm}                           & Joint only & 78.7  & 84.9	  & 68.7  & 69.6 & 88.8 & 48.3 \\
      CMD  \cite{cmd}                           & Joint only & 79.4  & 86.9	 & 70.3  & 71.5 & - & 43.0  \\
      \midrule[0.4pt]
      SkeletonMAE* \cite{skeletonmae}        & Joint only & 74.8  & 77.7	 & 72.5  & 73.5 & 82.8 & 36.1 \\
      \textbf{MAMP (Ours)}                      & Joint only & \textbf{84.9}  & \textbf{89.1}	 & \textbf{78.6}  & \textbf{79.1} & \textbf{92.2} & \textbf{53.8} \\
      \bottomrule[1.0pt]
    \end{tabular}
	\vspace{-2mm}
	\caption{Performance comparison on the NTU-60, NTU-120, and PKU-MMD datasets under the linear evaluation protocol. * indicates the re-implemented version under our framework, where improved performance is achieved.}
	\label{tab:sota_comp_linprobe}
    \vspace{-3mm}
\end{table*}

\begin{table}[t!]
	\centering
    \setlength\tabcolsep{4.3pt}
    \begin{tabular}{lccc}
      \toprule[0.8pt]
      \multirow{2}{*}{Method} & \multirow{2}{*}{Backbone} & \multicolumn{2}{c}{\textbf{NTU-60}}\\
      \cmidrule[0.4pt](lr){3-4}
                                                & & X-sub & X-view \\
      \midrule[0.4pt]
      CPM \cite{cpm}                            & ST-GCN      & 84.8    & 91.1 \\
      CrosSCLR \cite{li20213d}                  & 3s-ST-GCN   & 86.2    & 92.5 \\
      AimCLR \cite{guo2022aimclr}               & 3s-ST-GCN   & 86.9    & 92.8 \\
      CrosSCLR \cite{li20213d}                  & STTFormer   & 84.6    & 90.5 \\
      AimCLR \cite{guo2022aimclr}               & STTFormer   & 83.9    & 90.4 \\
      SkeletonMAE \cite{skeletonmae}            & STTFormer   & 86.6    & 92.9 \\
      Colorization \cite{yang2021skeleton}      & 3s-DGCNN    & 88.0    & 94.9 \\
      MCC \cite{Su_2021_ICCV}                   & 2s-AGCN     & 89.7    & 96.3 \\
      ViA \cite{yang2022via}                    & 2s-UINK     & 89.6    & 96.4 \\
      Hi-TRS \cite{chen2022hierarchically}      & 3s-Transformer & 90.0 & 95.7 \\
      \midrule[0.4pt]
      W/o pre-training                          & Transformer & 83.1    & 92.6 \\
      SkeletonMAE* \cite{skeletonmae}        & Transformer & 88.5    & 94.7 \\
      \textbf{MAMP (Ours)}                      & Transformer & \textbf{93.1}  & \textbf{97.5} \\
      \bottomrule[0.8pt]
    \end{tabular}
    \vspace{-2mm}
    \caption{Performance comparison on the NTU-60 dataset under the fine-tuned evaluation protocol.}
    \label{tab:sota_comp_finetune_ntu60}
\end{table}

\begin{table}[t!]
	\centering
    \setlength\tabcolsep{5.3pt}
    \begin{tabular}{lccc}
      \toprule[0.8pt]
      \multirow{2}{*}{Method} & \multirow{2}{*}{Backbone} & \multicolumn{2}{c}{\textbf{NTU-120}}\\
      \cmidrule[0.4pt](lr){3-4}
                                                & & X-sub & X-set \\
      \midrule[0.4pt]
      CPM \cite{cpm}                            & ST-GCN      & 78.4  & 78.9 \\
      CrosSCLR \cite{li20213d}                  & 3s-ST-GCN   & 80.5  & 80.4 \\
      AimCLR \cite{guo2022aimclr}               & 3s-ST-GCN   & 80.1  & 80.9 \\
      CrosSCLR \cite{li20213d}                  & STTFormer   & 75.0  & 77.9 \\
      AimCLR \cite{guo2022aimclr}               & STTFormer   & 74.6  & 77.2 \\
      SkeletonMAE \cite{skeletonmae}            & STTFormer   & 76.8  & 79.1 \\
      MCC \cite{Su_2021_ICCV}                   & 2s-AGCN     & 81.3  & 83.3 \\
      ViA \cite{yang2022via}                    & 2s-UINK     & 85.0  & 86.5 \\
      Hi-TRS \cite{chen2022hierarchically}      & 3s-Transformer & 85.3 & 87.4 \\
      \midrule[0.4pt]
      W/o pre-training                          & Transformer & 76.8  & 79.7 \\
      SkeletonMAE* \cite{skeletonmae}        & Transformer & 87.0  & 88.9 \\
      \textbf{MAMP (Ours)}                      & Transformer & \textbf{90.0}  & \textbf{91.3} \\
      \bottomrule[0.8pt]
    \end{tabular}
    \vspace{-2mm}
    \caption{Performance comparison on the NTU-120 dataset under the fine-tuned evaluation protocol.}
    \label{tab:sota_comp_finetune_ntu120}
\end{table}

\begin{table}[h]
    \setlength\tabcolsep{6.5pt}
    \centering
    \begin{tabular}{lcccc}
      \toprule
      \multirow{2}{*}{Method} & \multicolumn{2}{c}{\textbf{NTU-60}} & \multicolumn{2}{c}{\textbf{NTU-120}}\\
                                           \cmidrule[0.4pt](lr){2-3} \cmidrule[0.4pt](lr){4-5}
                                          & X-sub & X-view & X-sub & X-set \\
      \midrule
      PoseC3D \cite{duan2022revisiting}   & \textbf{93.7}  & 96.6 & 86.0 & 89.6 \\
      CTR-CGN \cite{Chen_2021_ICCV}      & 92.4  & 96.8 & 88.9 & 90.6 \\
      InfoGCN \cite{chi2022infogcn}       & 93.0  & 97.1 & 89.8 & 91.2 \\
      \midrule
      \textbf{MAMP (Ours)}   & 93.1  & \textbf{97.5} & \textbf{90.0}  & \textbf{91.3} \\
      \bottomrule
    \end{tabular}
    \vspace{-2mm}
    \caption{Performance comparison with fully-supervised methods on the NTU-60 and NTU-120 datasets. Note that our MAMP does not perform multi-stream ensembling during evaluation.}
    \label{tab:sota_comp_finetune_supervised}
\end{table}

\subsection{Experimental Setup}

\parsection{Network Architecture}
In our MAMP framework, we adopt a vanilla vision transformer \cite{dosovitskiy2020vit} as the backbone network, which consists of $L_e=8$ identical building blocks. In each block, the embedding dimension is 256, the head number of the multi-head self-attention module is 8, and the hidden dimension of the feed-forward network is 1024. We employ learnable spatio-temporal separated positional embedding to the embedded inputs. The settings of the transformer decoder used during pre-training are consistent with those of the backbone encoder except that the number of layers $L_d$ is reduced to 5.

\parsection{Data Processing Details} Given an original skeleton sequence, a contiguous segment is first randomly cropped from it with a certain proportion $p$ ($p$ is sampled from [0.5,1] during training and fixed to 0.9 during the test). After that, the cropped segment is resized to a fixed length $T_s$ by bilinear interpolation. $T_s$ is set to 120 by default.

\parsection{Pre-training Details} During pre-training, the masking ratio of the input token is set to 90\%. The target motion sequence $M^{\text{target}}$ has stride $m = 1$ and is padded with zeros. We adopt the AdamW \cite{loshchilov2018decoupled} optimizer with a weight decay of 0.05 and betas of (0.9, 0.95). We pre-train the network for 400 epochs with a batch size of 128. The learning rate is linearly increased to 1e-3 from 0 in the first 20 warm-up epochs and then decreased to 5e-4 by the cosine decay schedule. The experiments are conducted using the PyTorch framework on four NVIDIA RTX 3090 GPUs.

\subsection{Comparison with State-of-the-art Methods}

\parsection{Linear Evaluation Results}
In linear evaluation protocol, the pre-trained backbone is fixed and a post-attached linear classifier is trained with supervision for 100 epochs with a batch size of 256 and a learning rate of 0.1. The learning rate is decreased to 0 by the cosine decay schedule. As shown in Table \ref{tab:sota_comp_linprobe}, the performance on three datasets are reported, they are NTU-60, NTU-120, and PKU-MMD, respectively. We include latest high-performance approaches for comparison, \emph{e.g.}, GL-Transformer \cite{gl_transformer}, CPM \cite{cpm}, CMD \cite{cmd}, 3s-CrosSLR \cite{li20213d}, and 3s-AimCLR \cite{guo2022aimclr}. As we can see, with the joint stream as the only input, our proposed MAMP outperforms these methods on all the datasets. Specifically, MAMP outperforms previous state-of-the-art method CMD by 5.5\% and 8.3\% on the challenging NTU-60 x-sub and NTU-120 x-sub, respectively. For a fair comparison, we also re-implement the SkeletonMAE \cite{skeletonmae} under the same settings as our approach  (denote as SkeletonMAE*), where improved performance is achieved. We can find that our MAMP significantly outperforms SkeletonMAE* on all six subsets of the three datasets, demonstrating the superiority of masked motion prediction compared to the self-reconstruction of joints.

\parsection{Fine-tuned Evaluation Results}
In fine-tuned evaluation protocol, an MLP head is attached to the pre-trained backbone and the whole network is fully fine-tuned for 100 epochs with a batch size of 48. The learning rate is linearly increased to 3e-4 from 0 in the first 5 warm-up epochs and then decreased to 1e-5 by the cosine decay schedule. We also adopt layer-wise lr decay \cite{clark2020electra} following \cite{bao2022beit}. As shown in Table \ref{tab:sota_comp_finetune_ntu60} and Table \ref{tab:sota_comp_finetune_ntu120}, we evaluate the fine-tuned performance on NTU-60 and NTU-120, respectively. The vanilla transformer does not show satisfactory performance when trained directly from scratch, which is under expectation as weakly biased transformers require a large amount of training data to effectively prevent overfitting. After being pre-trained with the proposed MAMP framework, the network exhibits significant performance improvements ranging from 5\% to 13\% on the four subsets of the NTU-60 and NTU-120 datasets. The final results exceed all previous methods, even those with multi-stream ensembling such as Colorization \cite{yang2021skeleton}, MCC \cite{Su_2021_ICCV}, and ViA \cite{yang2022via}. Moreover, our MAMP also outperforms the re-implemented SkeletonMAE* by a considerable margin.

We also compare our MAMP with the top-performing supervised methods like PoseC3D \cite{duan2022revisiting}, CTR-GCN \cite{Chen_2021_ICCV}, and InfoGCN \cite{chi2022infogcn} in Table~\ref{tab:sota_comp_finetune_supervised}. Results show that without ensembling, MAMP outperforms most top-performing methods, especially on the larger NTU-120 dataset.

\parsection{Semi-supervised Evaluation Results} Following previous works \cite{li20213d,cmd,thoker2021skeleton}, in semi-supervised evaluation protocol, the post-attached classification layer and the pre-trained encoder are fine-tuned together with only a small fraction of the training set. Apart from that, we keep other training settings consistent with the fine-tuned evaluation protocol. As in \cite{li20213d,guo2022aimclr,cpm}, we report the performance on the NTU-60 dataset when using 1\% and 10\% of the training set. Note that considering the randomness during training data selection, we report the average of five runs as the final results. As shown in Table \ref{tab:sota_comp_semi}, our proposed MAMP significantly outperforms previous works like 3s-AimCLR \cite{guo2022aimclr}, CPM \cite{cpm}, CMD \cite{cmd}, and SkeletonMAE* \cite{skeletonmae}. When using only 1\% of the training data, MAMP outperforms SkeletonMAE* by 11.6\% and 14.0\% in X-sub and X-view respectively. Compared to training from scratch, MAMP pre-training brings performance improvements of more than 15.5\% on all subsets of NTU-60.

\parsection{Transfer Learning Evaluation Results} In transfer learning evaluation protocol, the network is pre-trained on a source dataset and then finetuned on a different target dataset. In this way, the generalizability of the learned representation is verified. In this paper, the target dataset is PKU-MMD II and the source datasets are NTU-60, NTU-120, and PKU-MMD I, respectively. Results in Table \ref{tab:sota_comp_transfer} show that, compared to previous methods, the representation learned by our proposed MAMP framework exhibits the best transferability, outperforming the reproduced SkeletonMAE* \cite{skeletonmae} by 12.2\%, 12.2\%, and 7.6\% on the three source datasets, respectively.

\begin{table}[t!]
	\centering
    \setlength\tabcolsep{6pt}
	\begin{tabular}{lcccc}
      \toprule[0.8pt]
      \multirow{3}{*}{Method} & \multicolumn{4}{c}{\textbf{NTU-60}}\\
      \cmidrule[0.4pt](lr){2-5}
      & \multicolumn{2}{c}{X-sub} & \multicolumn{2}{c}{X-view} \\
      \cmidrule[0.4pt](lr){2-3} \cmidrule[0.4pt](lr){4-5}
      & (1\%) & (10\%) & (1\%) & (10\%) \\
      \midrule[0.4pt]
      LongT GAN \cite{zheng2018unsupervised}    & 35.2 & 62.0 &  -    & -    \\
      MS$^2$L \cite{lin_ms2l_mm}                & 33.1 & 65.1 &  -    & -    \\
      ASSL \cite{si2020adversarial}             & -    & 64.3 &  -    & 69.8 \\
      ISC \cite{thoker2021skeleton}             & 35.7 & 65.9 &  38.1 & 72.5 \\
      3s-CrosSCLR \cite{li20213d}               & 51.1 & 74.4 &  50.0 & 77.8 \\
      3s-Colorization \cite{yang2021skeleton}   & 48.3 & 71.7 &  52.5 & 78.9 \\
      CMD \cite{cmd}                            & 50.6 & 75.4 &  53.0 & 80.2 \\
      3s-Hi-TRS \cite{chen2022hierarchically}   & 49.3 & 77.7 &  51.5 & 81.1 \\
      3s-AimCLR \cite{guo2022aimclr}            & 54.8 & 78.2 &  54.3 & 81.6 \\
      3s-CMD \cite{cmd}                         & 55.6 & 79.0 &  55.5 & 82.4 \\
      CPM \cite{cpm}                            & 56.7 & 73.0 &  57.5 & 77.1 \\
      \midrule[0.4pt]
      W/o pre-training                          & 38.8 & 70.8 & 40.4  & 76.0 \\
      SkeletonMAE* \cite{skeletonmae}        & 54.4 & 80.6 & 54.6  & 83.5 \\
      \textbf{MAMP (Ours)}                      & \textbf{66.0} & \textbf{88.0} & \textbf{68.7}  & \textbf{91.5} \\
      \bottomrule[0.8pt]
    \end{tabular}
	\vspace{-2mm}
	\caption{Performance comparison on the NTU-60 dataset under the semi-supervised evaluation protocol. We report the average of five runs as the final performance.}
	\label{tab:sota_comp_semi}
\end{table}

\begin{table}[t!]
	\centering
    \setlength\tabcolsep{6.4pt}
    \begin{tabular}{lccc}
      \toprule[0.8pt]
      \multirow{2}{*}{Method}   & \multicolumn{3}{c}{\textbf{To PKU-II}} \\
      \cmidrule[0.4pt](lr){2-4}
      & NTU-60  & NTU-120  & PKU-I \\
      \midrule[0.4pt]
      LongT GAN \cite{zheng2018unsupervised}  & 44.8  & - & 43.6  \\
      MS$^2$L \cite{lin_ms2l_mm}     		  & 45.8  & - & 44.1  \\
      ISC \cite{thoker2021skeleton}    		  & 51.1  & 52.3 & 45.1  \\
      CMD \cite{cmd}                          & 56.0  & 57.0 & -     \\
      \midrule[0.4pt]
      SkeletonMAE* \cite{skeletonmae}      & 58.4  & 61.0 & 62.5  \\
      \textbf{MAMP (Ours)}                    & \textbf{70.6}  & \textbf{73.2} & \textbf{70.1}  \\
      \bottomrule[0.8pt]
    \end{tabular}
	\vspace{-2mm}
    \caption{Performance comparison on the PKU-II dataset under the transfer learning evaluation protocol. The source datasets are the NTU-60, NTU-120, and PKU-I datasets.}
	\label{tab:sota_comp_transfer}
\end{table}

\begin{figure*}[t!]
    \centering
    \begin{minipage}{.31\linewidth}
        \centering
        \includegraphics[width=1.0\linewidth]{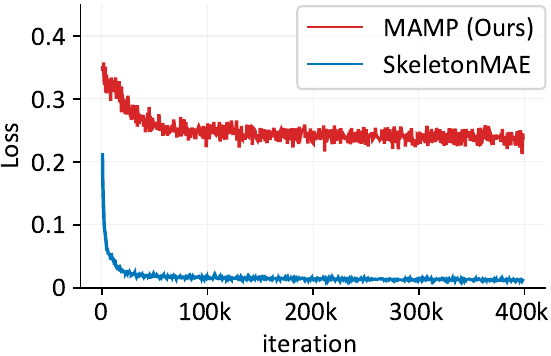}
        \vspace{-6mm}
        \caption{Pre-training loss plot. Compared to masked self-reconstruction of joints in SkeletonMAE, masked motion prediction acts as a harder pre-training objective.}
        \label{fig:loss}
    \end{minipage}
    \hspace{2mm}
    \begin{minipage}{.31\linewidth}
        \centering
        \includegraphics[width=1.0\linewidth]{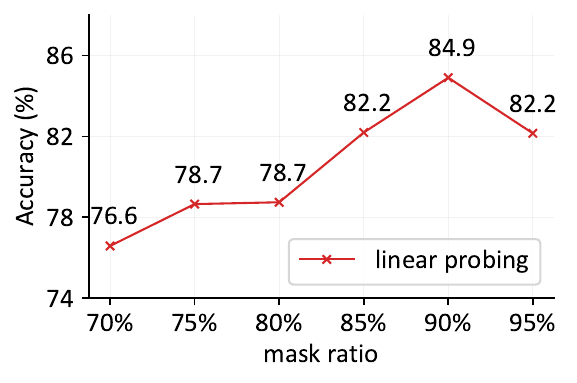}
    	\vspace{-6mm}
    	\caption{Ablation study on the masking ratio. The performance is evaluated on the NTU-60 X-sub dataset under the linear evaluation protocol.}
    	\label{fig:mask_ratio}
    \end{minipage}
    \hspace{2mm}
    \begin{minipage}{.31\linewidth}
          \centering
        \includegraphics[width=1.0\linewidth]{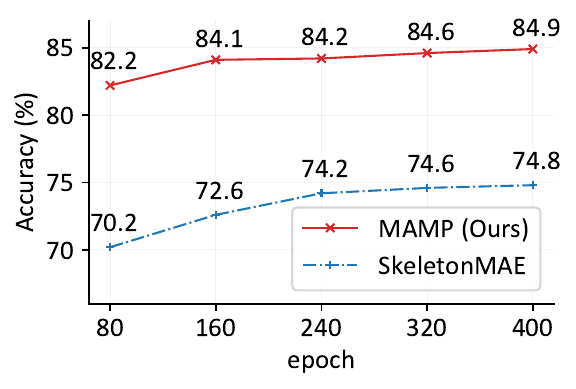}
        \vspace{-6mm}
        \caption{Pre-training schedule of SkeletonMAE and our MAMP. The performance is evaluated on the NTU-60 X-sub dataset under the linear evaluation protocol.}
        \label{fig:schedule}
    \end{minipage}
\end{figure*}

\subsection{Ablation Study}

\parsection{Superiority of Masked Motion Prediction}
As shown in Table \ref{tab:ablation_motion_prediction}, to verify the superiority of masked motion prediction, we designed four different ablative experiments. Given the joint and motion streams of the original data, all possible choices of the model input and reconstruction target are traversed. We can find that our joint-to-motion prediction significantly outperforms other strategies. Under the linear evaluation protocol, our MAMP exceeds joint-to-joint prediction (adopted by \cite{skeletonmae}) by 10.1\% and 6.1\% on NTU-60 and NTU-120, respectively. This suggests that predicting dynamic motion from static skeletons during pre-training facilitates better contextual modeling of 3D human actions.

We also visualize the pre-training loss in Figure~\ref{fig:loss}. Unlike the fast convergence of SkeletonMAE, masked motion prediction serves as a much harder pre-training objective.

\parsection{Mask Sampling Strategy}
In our approach, we employ the vanilla transformer as the backbone network, where embedding features at any spatio-temporal location can be freely masked as in MAE \cite{he2022masked}. To verify the effectiveness of the proposed motion-aware masking strategy, we compare its performance with that of random masking in Table \ref{tab:ablation_mask_sampling}. As we can see, our motion-aware masking strategy brings absolute performance improvements of 1.2\% and 1.3\% on NTU-60 and NTU-120, respectively. This suggests that the motion information, as an empirical semantic richness prior, can effectively guide the skeleton masking process.

\begin{table}[t!]
	\centering
    \setlength\tabcolsep{10pt}
    \begin{tabular}{cccc}
      \toprule[0.8pt]
      Input   & Target  & NTU-60  & NTU-120 \\
      \midrule[0.4pt]
      \multirow{2}{*}{Joint}  & Joint   & 74.8 & 72.5 \\
                              & Motion  & \baseline{\textbf{84.9}} & \baseline{\textbf{78.6}} \\
      \midrule[0.4pt]
      \multirow{2}{*}{Motion} & Joint   & 76.5 & 71.0 \\
                              & Motion  & 75.9 & 70.5 \\
      \bottomrule[0.8pt]
    \end{tabular}
	\vspace{-2mm}
    \caption{Ablation study on the superiority of masked motion prediction. The performance is evaluated on the NTU-60 X-sub and NTU-120 X-sub datasets under the linear evaluation protocol.}
	\label{tab:ablation_motion_prediction}
\end{table}

\begin{table}[t!]
	\centering
    \setlength\tabcolsep{8pt}
    \begin{tabular}{ccc}
      \toprule[0.8pt]
      Strategy     & NTU-60  & NTU-120 \\
      \midrule[0.4pt]
      Random masking        & 83.7 & 77.3 \\
      Motion-aware masking  & \baseline{\textbf{84.9}} & \baseline{\textbf{78.6}} \\
      \bottomrule[0.8pt]
    \end{tabular}
	\vspace{-2mm}
  \caption{Ablation study on the mask sampling strategy. The performance is evaluated on the NTU-60 X-sub and NTU-120 X-sub datasets under the linear evaluation protocol.}
  \label{tab:ablation_mask_sampling}
\end{table}

\begin{table}[t!]
    \centering
    \setlength\tabcolsep{6pt}
    \begin{tabular}{ccccc}
      \toprule[0.8pt]
      $T_e=T_s / l$ &   $T_s$   &   $l$\quad   & NTU-60  & NTU-120 \\
      \midrule[0.4pt]
      \multirow{4}{*}{30}    &   60      &   2\quad   & 84.6 & 77.8 \\
          &   120     &   4\quad   & \baseline{\textbf{84.9}} & \baseline{\textbf{78.6}} \\
          &   180     &   6\quad   & 83.3 & 78.6 \\
          &   240     &   8\quad   & 84.5 & 77.9 \\
      \bottomrule[0.8pt]
    \end{tabular}
    \vspace{-2mm} 
    \caption{Ablation study on the segment length $l$ used in the joint embedding process. For a fair comparison, the input length $T_s$ is adjusted to ensure that the embedded features have a fixed length $T_e$. The performance is evaluated on the NTU-60 X-sub and NTU-120 X-sub datasets under the linear evaluation protocol.}
    \label{tab:ablation_segment}
\end{table}

\begin{table}[t!]
  \centering
  \begin{minipage}{.49\linewidth}
    \centering
    \setlength\tabcolsep{3pt}
    \begin{tabular}{ccc}
      \toprule[0.8pt]
      $L_d$     & NTU-60  & NTU-120 \\
      \midrule[0.4pt]
      2   & 83.3 & 77.2 \\
      3   & \baseline{\textbf{84.9}} & 77.6 \\
      4   & 84.6 & 77.5 \\
      5   & \baseline{\textbf{84.9}} & \baseline{\textbf{78.6}} \\
      \bottomrule[0.8pt]
    \end{tabular}
    \subcaption{Decoder depth.}
  \end{minipage}
  \begin{minipage}{.49\linewidth}
    \centering
    \setlength\tabcolsep{3pt}
    \begin{tabular}{ccc}
      \toprule[0.8pt]
      $C_d$ & NTU-60  & NTU-120 \\
      \midrule[0.4pt]
      64    & 82.3 & 74.2 \\
      128   & 83.5 & 77.3 \\
      256   & \baseline{\textbf{84.9}} & \baseline{\textbf{78.6}} \\
      512   & 84.5 & 77.1 \\
      \bottomrule[0.8pt]
    \end{tabular}
    \subcaption{Decoder width.}
  \end{minipage}
  \vspace{-2mm}
  \caption{Ablation study on the decoder design. The performance is evaluated on the NTU-60 X-sub and NTU-120 X-sub datasets under the linear evaluation protocol.}
  \label{tab:ablation_decoder}
\end{table}

\parsection{Segment Length}
We evaluated the performance of the learned representation under different segment lengths $l$ used in the joint embedding process. For a fair comparison, we resize the original input sequence to ensure that the embedded features have a fixed length $T_e = T_s / l = 30$. As shown in Table \ref{tab:ablation_segment}, a segment length of 4 brings the best performance on both NTU-60 and NTU-120 datasets.

\parsection{Decoder Design}
We experimented with different numbers of layers and widths (feature dimensions) for the transformer decoder. As shown in Table \ref{tab:ablation_decoder} (a), our MAMP exhibits the best performance on the NTU-60 and NTU-120 datasets when the number of decoder layers is 3 and 5, respectively. The experimental results of the decoder width are in Table \ref{tab:ablation_decoder} (b), where a width of 256 brings the best performance. Overall, a decoder with 5 layers and a width of 256 is adopted in our MAMP framework by default.

\parsection{Masking Ratio}
As shown in Figure~\ref{fig:mask_ratio}, we experimented with different masking ratios. Results on NTU-60 X-sub show that either too large or too small masking ratios have a negative impact on performance. We empirically found that a masking ratio of 90\% exhibits the best results.

\parsection{Pre-training Schedule} We studied the influence of the pre-training schedule length. As shown in Figure \ref{fig:schedule}, both SkeletonMAE and our MAMP exhibit higher performance with longer pre-training schedules. It is worth mentioning that our MAMP significantly outperforms SkeletonMAE for all pre-training schedules with lengths ranging from 80 to 400 epochs, demonstrating the stability and superiority of the proposed masked motion prediction strategy.

\section{Conclusion}
In this work, we present MAMP, a simple yet effective framework for 3D action representation learning. We show that compared to conventional masked self-reconstruction of human joints, masked joint-to-motion prediction is demonstrated to be more effective for contextual motion modeling of 3D human actions. Given the high temporal redundancy of the skeleton sequence, we further devise the motion-aware masking strategy, which incorporates motion intensity as the empirical semantic richness prior for adaptative skeleton masking, facilitating better attention to semantically rich temporal regions. We conduct extensive experiments on three prevalent benchmarks under four evaluation protocols. Results show that the proposed MAMP brings remarkable performance improvements and sets a series of new state-of-the-art records, unleashing the tremendous potential of vanilla transformers for 3D action modeling.

~\\
\parsection{Acknowledgement} This work was supported by NSFC under Contract U20A20183 \& 61836011. It was also supported by the GPU cluster built by MCC Lab of Information Science and Technology Institution, USTC, and the Supercomputing Center of the USTC.

{\small
\bibliographystyle{ieee_fullname}
\bibliography{egbib}
}

\end{document}